\documentclass[11pt, letterpaper]{article}

\usepackage{microtype}
\usepackage{hyperref}
\usepackage{url}
\usepackage{booktabs}

\usepackage{lineno}

\usepackage{array}
\usepackage{multirow}

\usepackage{authblk}
\usepackage{adjustbox}

\usepackage{xspace}

\usepackage[utf8]{inputenc}
\usepackage[T1]{fontenc}
\usepackage{amsmath}
\usepackage{amsfonts}
\usepackage{amssymb}
\usepackage{amsthm}
\usepackage{mathtools}
\usepackage{tabularx}
\usepackage{enumerate}
\usepackage{graphicx}
\usepackage[left=1in,right=1in,top=1in,bottom=1in]{geometry}
\usepackage{hyperref}
\usepackage{natbib}

\usepackage[english]{babel}

\usepackage{bbm}
\usepackage{tcolorbox}
\usepackage[normalem]{ulem}

\usepackage{microtype}
\usepackage{subfigure}
\usepackage{booktabs}
\usepackage{float}
\usepackage{algorithm}
\usepackage{algorithmic}
\usepackage{appendix}
\newcommand\blfootnote[1]{%
  \begingroup
  \renewcommand\thefootnote{}\footnotetext{#1}%
  \endgroup
}

\usepackage[capitalize,noabbrev]{cleveref}

\newcommand{\tss}{\tilde{S}}

\newcommand{\Var}{\mathop{\mathrm{Var}}}

\newcommand{\llmaaj}{LLM-as-a-judge\xspace}

\allowdisplaybreaks

\title{Play Favorites: A Statistical Method \\to Measure Self-Bias in LLM-as-a-Judge}

\author{Evangelia Spiliopoulou$^\dag$}
\author{Riccardo Fogliato$^\dag$}
\author{Hanna Burnsky}
\author{Tamer Soliman}
\author{Jie Ma}
\author{Graham Horwood}
\author{Miguel Ballesteros}
\affil{Amazon Web Services}

\begin{document}

\maketitle

\begin{abstract}
Large language models (LLMs) can serve as judges that offer rapid and reliable assessments of other LLM outputs. However, models may systematically assign overly favorable ratings to their own outputs—a phenomenon known as \emph{self-bias}—which can distort evaluations of true model performance. Previous studies often conflate genuine differences in model quality with bias or incorrectly assume that evaluations from LLMs and humans follow the same rating distributions. In this work, we present a statistical framework that explicitly formalizes assumptions under which self-bias can be identified and estimated. Our method models the difference in the scoring distribution that \llmaaj assigns to its own completions compared to other models, while accounting for the underlying quality of the completions provided by an independent, third-party judge (e.g., humans). Our method reliably isolates and quantifies self-bias, even when models vary in ability, ensuring that genuine performance differences are not mistaken for self-bias. We conduct an empirical analysis of self-bias on a large dataset (>5000 prompt-completion pairs) consisting of expert human annotations and judgments from nine different LLM judges$^\ast$. We find that some models, such as GPT-4o and Claude 3.5 Sonnet, systematically assign higher scores to their own outputs. These models also display \emph{family-bias}; systematically assigning higher ratings to outputs produced by other models of the same family. Our findings highlight potential pitfalls of using LLM judges and offer practical guidance to mitigate biases when interpreting automated evaluations.

\end{abstract}

\blfootnote{$^\dag$Equal contribution.}
\blfootnote{$^\ast$Code and Data: https://github.com/spilioeve/Play-Favorites}

\section{Introduction}

With the ever-growing abilities of large language models (LLMs), there is an increasing demand for more tailored and reference-free evaluation than traditional NLP metrics \citep{lin2004rouge,papineni2002bleu,snover2006study}. LLMs are increasingly adopted as evaluators to judge the quality of outputs generated by other models \citep{zheng2023judging,liu2023g,chiang2023can}. However, LLM-as-judges are shown to exhibit several types of
biases, such as positional bias, self-enhancement bias, and verbosity bias, among others \citep{zheng2023judging,wang2024large,liu2023llms}. 
In this work, we focus specifically on self-enhancement bias, also known as \emph{self-bias}. Informally, self-bias occurs when an LLM-as-a-judge systematically assigns higher scores to its own outputs compared to equally good outputs from other models, as scored by a reliable independent judge (e.g., a human expert).

Prior work on self-bias can be categorized into two main directions. One direction compares how an \llmaaj scores multiple models, concluding that self-bias exists if a judge systematically assigns higher scores to its own outputs \citep{panickssery2024llm}. Yet, this may mistakenly attribute high scores to bias, even when the \llmaaj genuinely produces higher-quality completions than the other evaluated models. Another direction contrasts \llmaaj scores of its own completions with those of an independent judge (such as a human) \citep{xu2024pride, wataoka2024self}. However, this approach fails to account for consistent annotation differences between two judges (e.g., a judge may be consistently more lenient than the other). While recent efforts have attempted to integrate these two approaches \citep{liu2023llms}, they do not provide a formal statistical framework with clear assumptions and criteria for measuring self-bias. %

\paragraph{Our contributions} Via our work, we make three main contributions. First, we introduce a principled statistical framework to identify and quantify self-bias in \llmaaj that does not suffer from the above limitations and clearly specifies the assumptions required for valid inference (\Cref{sec:methods}). Our approach builds upon prior methods that compare each LLM's self-scores to scores from an independent judge, but employs a regression model that explicitly accounts for systematic differences between judges and enables formal statistical testing of self-bias. Second, along with this publication, we release a new dataset containing expert human evaluations of completions from nine LLMs (including Llama 3, GPT, Mistral, and Claude models) to almost 600 prompts along six evaluation dimensions, with associated judgments from the same LLMs (\Cref{sec:data}). Third, we conduct an empirical analysis on this dataset, where we find evidence of positive self-bias and \emph{family-bias}, a tendency to favor completions from models within the same family (\Cref{sec:results}). 

\section{Related Work}

\subsection{Biases in LLM-as-a-judge Judgments}

Recent work on \llmaaj methods has expanded rapidly, as documented by \citet{li2024generation}, who survey hundreds of studies exploring diverse variants and applications. These methods differ in their core methodologies—ranging from detailed prompting strategies \citep{gao2023human,bai2022constitutional,ye2024justice} to models fine-tuned specifically for evaluation tasks \citep{wang2024pandalm,zhujudgelm,ligenerative,kim2024prometheus}—as well as in the evaluation attributes they target (e.g., faithfulness, relevance) and in their scoring mechanisms, either using single absolute scores per generation \citep{kocmi2023large} or pairwise comparisons that yield model rankings.

LLM judges are shown to exhibit systematic biases that favor completions with certain superficial characteristics rather than reflecting genuine quality differences. For example, \citet{wang2024large} focus on position bias in pairwise settings, where the relative order of evaluated outputs affects the scores, while \citet{zheng2023judging} report self-bias, verbosity, and position biases. Furthermore, \citet{stureborg2024large} find that \llmaaj tends to assign higher scores to completions with lower perplexity—a phenomenon often referred to as familiarity bias. Complementing these findings, \citet{park2024offsetbias} identify seven distinct bias types using a meta-evaluation framework based on hand-crafted test cases, and \citet{chen2024humans} demonstrate that biases such as misinformation oversight, gender, authority, and style bias are common in both \llmaaj and human evaluations. The CALM framework \citep{ye2024justice} quantifies multiple bias types by applying deliberate perturbations to mimic various characteristics. These results collectively underscore the need for rigorous statistical frameworks to identify, quantify, and mitigate bias in \llmaaj.

\subsection{Self-bias in \llmaaj}

Self-bias poses a significant challenge to the reliability of LLM evaluations \citep{deutsch2022limitations}. While other biases can be artificially introduced via perturbations of a completion, this is not the case for self-bias, making its study particularly challenging  \citep{zheng2023judging}. Research follows two methodological directions: either comparing \llmaaj scores across outputs of different LLMs and its own completions, or contrasting LLM self-scores to those of an independent third-party judge.

Within the first direction, \citet{koo2023benchmarking} and \citet{liu2023llms} examine the frequency at which an LLM assigns higher scores to their own outputs over others'. Similarly, \citet{panickssery2024llm} measure self-bias by using the difference of scores to its own completions to other models' scores, used to analyze the association between self-bias and self-recognition. However, these methods risk conflating self-bias with genuine performance differences.

Within the second direction, \citet{zheng2023judging} analyze win rates in pairwise comparisons between LLM and human evaluations on benchmarks like MT-Bench and Chatbot Arena. The authors interpret higher scores to its own completions as evidence of self-bias. Although this method may suggest self-bias, the authors do not dive into the statistical assumptions required in order to make a statistical inference on the presence and magnitude of the self-bias, e.g., how the two sets of scores are related. 

\citet{xu2024pride} propose a statistical framework under the assumption that there are no differences in rating distributions between LLM and human scores, however this assumption is not always correct.

\section{Methodology}\label{sec:methods}

\begin{figure}[t]
  \centering
  \includegraphics[width=\textwidth]{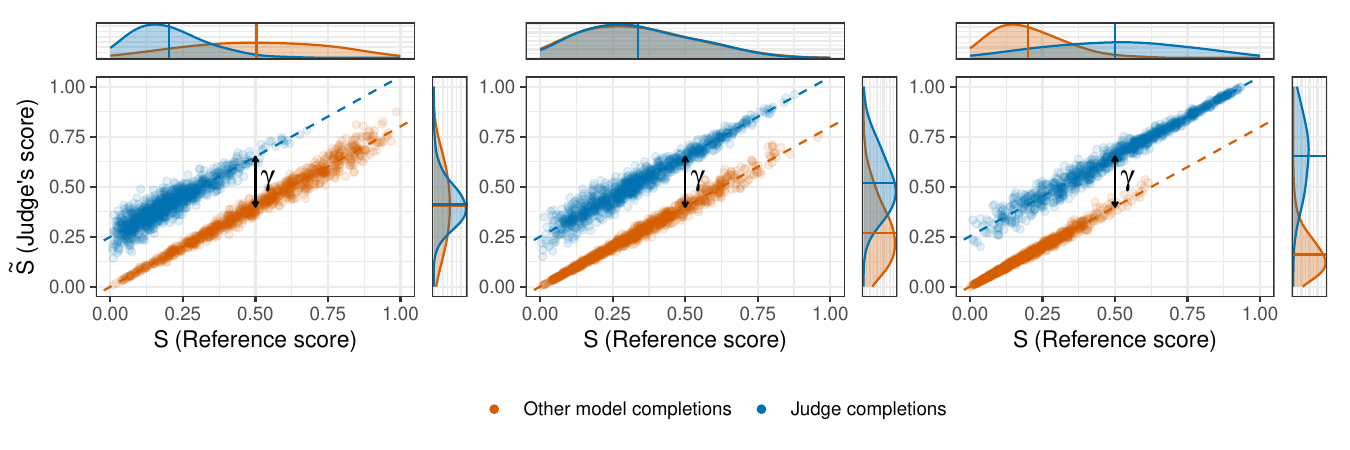}
  \caption{Illustration of our regression-based approach to measure self-bias, where
  $
  \tilde{S}_{imj} = \smash{\beta_j S_{im} + \gamma_j\,\mathbf{1}_j(m) + \epsilon_{imj}}
  $ with $\beta_j=0.8$ and $\gamma_j=0.25$. 
  Each main scatter plot displays the LLM-as-a-judge rating $\smash{\tilde{S}_{imj}}$ vs. the reference score $\smash{S_{im}}$, with regression lines for judge completions (offset by $\gamma_j$) and for other models no offset). For example, if the judge and another model happen to both have completions with the same quality $\smash{S_{im}=0.5}$, the judge would rate its own $\smash{\tilde{S}_{imj}=0.65}$ and $0.4$ the other model's. Side density plots display score distributions and mean values (vertical lines). Left: Judge completions are of lower quality, so self-bias partially compensates for the gap. Middle: Both groups have similar quality, making the self-bias more apparent. Right: Judge completions are of higher quality, and self-bias further increases the gap in the scores.}
  \label{fig:regression-visual}
\end{figure}

In this section, we introduce our approach for estimating self- and family-bias. Essentially, we compare how an \llmaaj rates its own completions vs. those from other models, while accounting for each completion’s underlying quality. Intuitively, if two completions have similar quality but the \llmaaj consistently scores its own higher, that discrepancy indicates self-bias. Since LLMs generate completions that may differ in quality, a direct comparison of the \llmaaj scores without controlling for this difference may not give reliable results of self-bias.

\paragraph{Notation}
Let $i=1,\ldots,N$ index prompts, $d=1,\ldots,D$ index evaluation dimensions, $m=1,\ldots,M$ index models that generate completions,
$j=1,\ldots,J$ index LLM judges.
For each prompt, every model produces a completion that is scored by the judges.
The set of LLMs that generate completions and act as judges need to partially or fully overlap in order to estimate self-bias. 
For prompt $i$, dimension $d$, and model $m$, denote
the \llmaaj rating
by $\tss_{idmj}\in\mathbb{R}$, for $j=1,\dots,J$.
We also have access to a reference score provided by a third-party judge denoted as $S_{idm}\in\mathbb{R}$, which will serve as our benchmark measures of completion quality.

\paragraph{Modeling approach}
We specify the following linear regression model for the rating $\tss_{idmj}$ assigned by the \llmaaj $j$:
\begin{align}\label{eq:main_model}
    \tss_{idmj}
     = \alpha
    + \underbrace{\delta_j + \beta_j\,S_{idm}}_{\text{Human alignment}}
    + \underbrace{\gamma_j\,\mathbf{1}_j(m)}_{\text{Self-bias}} 
     +  \underbrace{\lambda_{F(j)}\,\mathbf{1}_{F(j)}(F(m))}_{\text{Family-bias}} + \eta_d
    + \epsilon_{idmj}.
\end{align}
where $\alpha$ is a global intercept;  
$\beta_j$ is the judge-specific sensitivity to $S_{idm}$ and $\delta_j$ is a judge fixed effect. These terms account for the alignment between judge and reference scores.
To measure the favoritism of the judge with respect to its own completions,
we include the term $\gamma_j$, which measures the self-bias of model $m$ towards its own completions
(active when $j=m$). 
The coefficient $\lambda_{F(j)}$ captures family-bias, which is the favoritism of the
judge to models of the same family (active when $F(j)=F(m)$). 
To isolate self-bias from family favoritism, we set $\mathbf{1}_{F(j)}(F(j))=0$.
Note that in this model we pool data from all dimensions and include $\eta_d$, a dimension-specific fixed effect, to capture constant shifts in the judge vs. reference scores across dimensions.
Finally, $\epsilon_{idmj}$ is the classical error term. 

\paragraph{Interpreting the regression model}  
\Cref{fig:regression-visual} illustrates our regression-based approach using three simulated scenarios for a single judge, one evaluation dimension, and no family bias. In each panel, the main scatter plot (with accompanying side density plots) depicts the relationship between the judge’s ratings, \(\tilde{S}\), and the reference scores, \(S\). The regression slope, \(\beta\), quantifies the extent to which the judge’s ratings track the reference scores (with values near 1 indicating strong alignment), while the vertical offset, \(\gamma\), represents self-bias—that is, the extra boost the judge assigns to its own completions. In other words, if two completions have identical underlying quality \(S\), the judge’s own output is expected to be rated \(\gamma\) points higher. 

\paragraph{Why reference scores are necessary}  
Not using reference scores $S$ can lead to incorrect conclusions about self-bias. To better understand this, let's look at the simulations, where we fix the same self-bias $\gamma$ and alignment $\beta$, but vary the underlying quality distribution of the judge’s completions. In the left panel, the judge’s completions are of lower quality, so self-bias partly compensates for that gap, making scores of the judge appear deceptively similar. In the right panel, the judge’s completions have higher quality, causing the rating gap between judge and other completions to substantially exceed the actual self-bias. Only in the middle panel -- where both groups share similar quality -- does self-bias become clearly visible by comparing the judge’s scores alone. Hence, without reference scores $S$, one might underestimate or misinterpret the true magnitude of self-bias.

\paragraph{Estimating self- and family-bias} 
We estimate the coefficients of our regression model in \Cref{eq:main_model} using ordinary least squares (aka OLS). 
To determine whether self- and family-bias estimates are statistically significant, we quantify uncertainty around these estimates by computing 90\% Wald (Gaussian) confidence intervals using robust (White) standard errors \citep{buja2019models, cameron2015practitioner, freedman2006so}. 
This type of standard errors ensures valid inference even when certain model assumptions are violated. We then classify a coefficient as statistically significant at the 10\% level if its corresponding confidence interval does not contain zero (equivalently, if a Wald test rejects the null hypothesis); further details are provided in \Cref{sec:appendix-methods}.

\paragraph{Robustness checks}Checking that the measurements and conclusions drawn from the main modeling approach hold under different assumptions of the data-generating mechanism is a fundamental step in statistical analyses \citep{buja2019models2}. 
In \Cref{sec:results_robustness}, we conduct a series of robustness checks where we vary our model specifications (using generalized additive model and ordinal logit regression), control for the length of the completion, replace human scores with a third-party LLM scores, and analyze the self- and family-biases separately for each dimension and task.

\section{Data}\label{sec:data}

Our evaluation data consists of prompts sourced from publicly available datasets. Since many of the existing datasets contain completions generated by relatively weak LLMs \citep{helm-instruct,zheng2023judging}, we collect new model completions and corresponding \llmaaj judgments, which we will publicly release along with the publication. 
Here we present the key aspects of our data collection, with more details in \Cref{sec:appendix-human_quality}. The data is provided in the supplementary material.

\subsection{Prompts and Model Completions}

We use a set of 596 prompts selected from question-answering (QA) and summarization tasks previously employed in \llmaaj research \citep{panickssery2024llm,zheng2023judging}. Specifically, we include prompts from Chatbot Arena (139) and MT-Bench (53) \citep{zheng2023judging}, HELM-Instruct (160) \citep{helm-instruct}, Stanford Human Preferences (44) \citep{pmlr-v162-ethayarajh22a}, XSUM (100) \citep{narayan2018don}, and CNN/DailyMail (100) \citep{nallapati2016abstractive}. HELM-Instruct itself aggregates prompts from diverse sources \citep{bai2022training,perez2022red,geng2023koala,team2023vicuna,kopf2023openassistant,wang2023self,gridfiti2023}. From a larger initial pool, we select prompts that avoid potentially harmful or overly subjective requests (e.g., ``Tell me a joke''). For each chosen prompt, we generate completions using nine language models: Claude v2, Claude 3 Sonnet, Claude 3.5 Sonnet \citep{claude2023}; GPT-3.5 Turbo, GPT-4o \citep{achiam2023gpt,hurst2024gpt}; Llama 3 8B, Llama 3 70B \citep{grattafiori2024llama}; and Mistral 7B, Mistral Large \citep{jiang2023mistral7b}.

\subsection{Evaluation Dimensions}

\begin{table}
    \centering
    \begin{tabular}{m{0.25\linewidth} m{0.7\linewidth} }
        \toprule
        \textbf{Evaluation Dimension} & \textbf{Definition}  \\
        \midrule
        Completeness &
        Whether the output includes all needed information and details. \\
        Conciseness &
        Whether the output is focused on the input without irrelevant content. \\
        Logical robustness &
        Whether the reasoning in the output follows a clear flow. \\
        Logical correctness &
       Whether the output is factually accurate and addresses the input. \\
        Helpfulness &
        How useful and supportive the output is for most users. \\
        Faithfulness &
        Whether the output reflects input without adding unrelated information. \\
        \bottomrule
    \end{tabular}
    \caption{Evaluation dimensions based on which the quality of completions was assessed.}
    \label{tab:eval_dimensions}
\end{table}

We evaluate the resulting 5364 completions (596 prompts × 9 models) across six evaluation dimensions; see \Cref{tab:eval_dimensions} for their definitions. Each dimension is described in detail, with examples, in \Cref{sec:appendix-human_quality}. Specifically, we assess helpfulness, completeness, and conciseness using definitions from \citet{helm-instruct}; logical correctness and logical robustness using definitions from \citet{ye2024flask}; and faithfulness based on criteria from summarization tasks, where responses must accurately reflect the provided context \citep{maynez2020faithfulness}. Most dimensions are scored using a 5-point Likert scale, with exceptions for logical correctness (3-point scale) and helpfulness (7-point scale) to enable finer-grained distinctions. We use the same rubric for both human and \llmaaj scoring.

\subsection{Evaluation of Human Annotations}
\label{sec:human_eval}

Each prompt-completion pair is annotated by three human raters on every dimension, from an in-house team of annotators specifically trained with our evaluation guidelines. We aggregate the scores by taking the mean of the numerical scores for each example, which we use as reference score in our analysis of self-bias described in \Cref{sec:methods}. 

We assess the quality of these annotations in three ways: via their accuracy on an ``attention check'' set, the accuracy on a gold dataset (i.e., a random subset of 210 prompt-completion pairs, annotated by a separate team of annotators with more expertise and training on the specific guidelines), and the inter-annotator agreement. The attention check examples consist of simple perturbations on held-out prompt-completion pairs, that yield low scores on particular dimensions. 
During the annotation process, any annotator who repeatedly failed attention checks was removed from the task and their annotations were re-worked by other annotators.

Accuracy on the gold subset is over 84\% for all dimensions, with average 91\%, indicating that annotators have as good understanding of the guidelinesas the more experienced annotators. We additionally compute inter-annotator agreement on the entire  dataset. The average Krippendorff's $\alpha$ is 0.28 across all dimensions; however, as seen in \Cref{tab:quality}, some dimensions have significantly higher Krippendorff's $\alpha$, such as helpfulness and completeness with $\alpha$ = 0.47. Due to known problems with chance-corrected measures of inter-rater reliability when applied to datasets with highly skewed label distributions \citep{zhao2013assumptions}, we also estimate the observed agreement (i.e., how often all three annotators agree), which is high (81\%). See more details in \Cref{sec:appendix-human_quality}.

\subsection{Evaluation of LLM-as-a-Judge Scores}

We assess the quality of \llmaaj scores based on their correlation with human annotations for each evaluation dimension (see \Cref{fig:llmaaj_human_spearman}). Because the underlying data is ordinal, we use Spearman's tie-corrected rank correlation $\rho$ \citep{spearman1961proof}. Overall we observe higher correlation across dimensions for stronger models, such as GPT-4o and Claude-3.5-Sonnet. We further see a stronger correlation between \llmaaj and humans for dimensions with higher (human) inter-annotator agreement, such as completeness and helpfulness, where $\rho > 0.4 $. This indicates that the same dimensions are equally challenging for both humans and LLMs.

\section{Results}\label{sec:results}

We discuss our main results by starting with an exploratory analysis and then estimating self- and family-bias via our proposed approach (code in the supplementary material). We also conduct a brief analysis on HELM-Instruct data \citep{helm-instruct}, whose results can be found in \Cref{sec:helm-instruct-analysis}. 

\subsection{Exploratory Analysis}

\begin{figure}[t!]
    \centering
    \includegraphics[width=\textwidth]{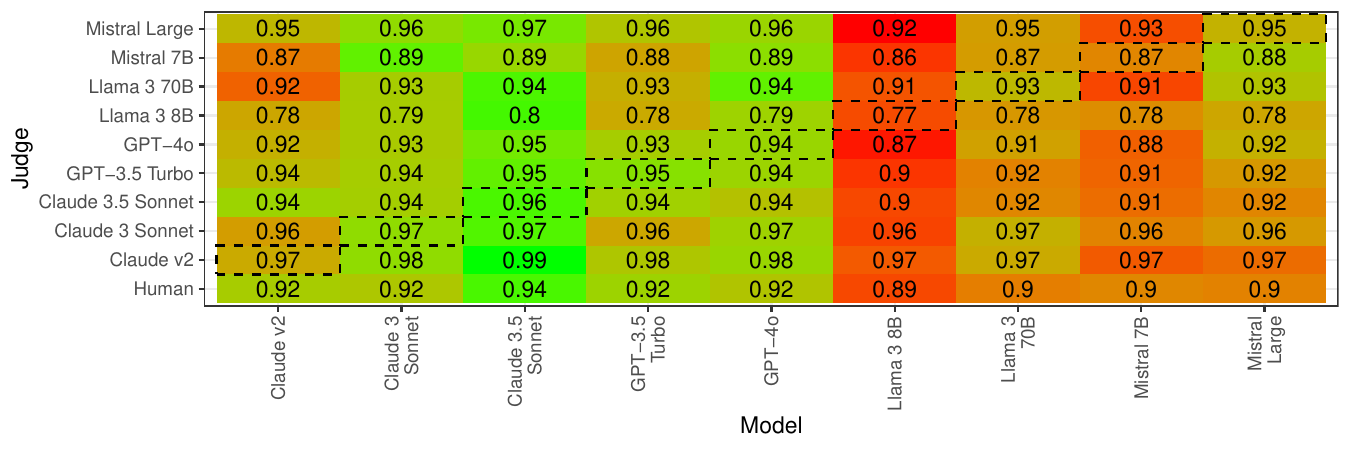}
    \caption{Heatmap of average LLM and human scores of LLM completions. LLM scores on their own completions are highlighted on the diagonal. Color scale is proportional to the average ratings normalized by row.}
    \label{fig:judge-and-human-scores-heatmap}
\end{figure}

Following prior work on self-bias \citep{liu2023llms}, we analyze how each \llmaaj evaluates its own completions compared to those of other models. \Cref{fig:judge-and-human-scores-heatmap} displays a heatmap summarizing the average scores each judge (rows) assigns to the outputs of each model (columns), averaged across evaluation dimensions. The scores mostly cluster within a narrow range (0.90–1.0), indicating that models generally rate each other's outputs positively and refrain from strong criticism. This behavior is expected due to the high-quality outputs generated by state-of-the-art models \citep{zheng2023judging}. The dashed diagonal cells highlight the scores models assign to their own completions.

While inspecting the diagonal cells row-wise or column-wise, some models (e.g., Claude-v2 or GPT-3.5-turbo) appear to assign higher scores to their own completions. However, there is no clear criterion of when such a pattern indicates self-bias and its extend.

Comparing how a judge scores its own outputs vs. others without accounting for the actual quality of the outputs (e.g., via human scores) risks falsely identifying self-bias whenever the other models produce lower-quality completions.

\begin{figure}[t]
    \centering
    \includegraphics[width=\textwidth]{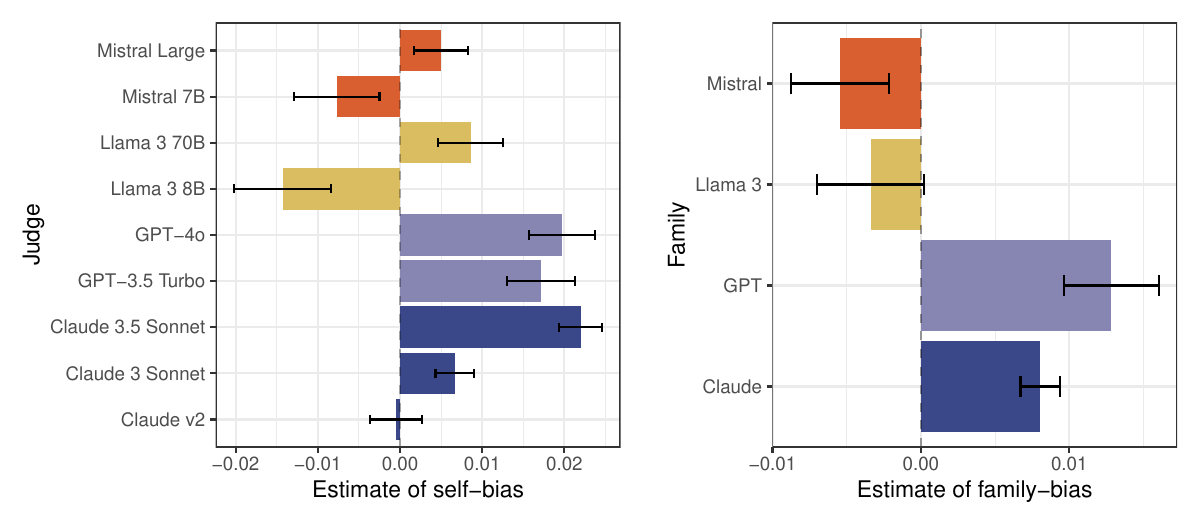}
    \caption{Estimates of self-bias ($\gamma_j$, left) and family-bias ($\lambda_{F(j)}$, right) with associated 90\% confidence intervals obtained using the approach described in \cref{sec:methods}, colored by the family.}
    \label{fig:self-and-family-bias}
\end{figure}

\subsection{Measuring Self- and Family-bias}\label{sec:self-bias}

\paragraph{Self-bias} To statistically estimate the magnitude of self-bias, we use our method from \Cref{sec:methods}. \Cref{fig:self-and-family-bias} shows the estimated self-bias coefficients for each \llmaaj ($\gamma_j$), along their 90\% confidence intervals. For the GPT models and Claude 3.5 Sonnet, we observe positive self-bias: a positive association between the completion being their own and higher scores, even after controlling for the quality of the completions. In contrast, weaker Claude models, such as Claude-v2 and Claude 3-Sonnet, exhibit almost no self-bias. Interestingly, Llama 3 8B displays significant negative self-bias. As we discuss in \Cref{sec:ablations}, self-bias may differ across evaluation dimensions (e.g., in Llama 3 8B), and, thus, compiling results across dimensions may not be representative of the model behavior.

\paragraph{Family-bias} Models that share architecture or training data might share a characteristic ``evaluation lens''. Thus, we evaluate family-bias, the tendency of models to favor outputs from other models within the same family. 
Again, we rely on the model in \Cref{sec:methods} and estimate $\lambda_{F(j)}$. 
As seen in \Cref{fig:self-and-family-bias}, we find that Claude and GPT judges tend to give higher scores to completions of other models within the same family. The tendency in both families is common across all models, e.g., Claude 3.5 Sonnet boosts the scores of both Claude v2 and of Claude 3 Sonnet. Llama and Mistral models do not exhibit such bias.

Although many of these effects may appear small, they can significantly impact model rankings, particularly because all models achieve high scores. For example, when comparing Claude Sonnet 3.5 and GPT-4o, a score difference of just 0.02 is comparable to the magnitude of the observed self-bias. While the practical significance of such shifts may depend on the application, it is important to be aware of these effects when interpreting evaluation results and making model comparisons.

\section{Analysis and Ablations}\label{sec:results_robustness}

Our analysis includes ablations of tasks and dimensions, as well as a series of robustness tests of the models considered and the different modeling approaches (e.g., inclusion of length bias). As the results show, the magnitude of self-bias varies across evaluation dimensions, but the trends of each model in \Cref{sec:results} are mostly consistent across all robustness checks.

\subsection{Slicing the Data}\label{sec:ablations} We analyze self- and family-bias by splitting the data in two ways and estimate a different linear model for each split: per evaluation dimension (e.g., faithfulness), and  per task type. 

\begin{figure}[t]
    \centering
    \includegraphics[width = \textwidth]{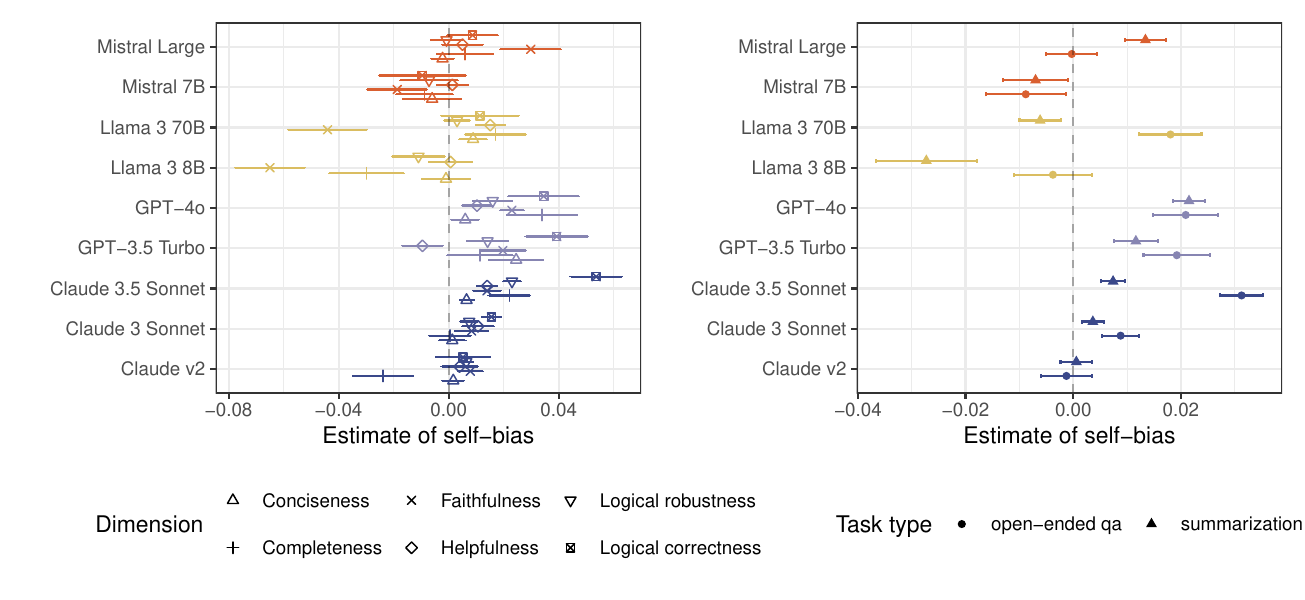}
    \caption{Estimates of self-bias ($\gamma_j$) obtained using the approach described in \Cref{sec:methods}, colored by the family, grouped by dimension (left) and by task type (right). Estimates are obtained by fitting the model in \Cref{eq:main_model} for each dimension or task separately.}
    \label{fig:bias-by-dimension-and-dataset}
\end{figure}

\paragraph{Analysis by evaluation dimension} We estimate the regression model from \Cref{eq:main_model} separately for each evaluation dimension. 

\Cref{fig:bias-by-dimension-and-dataset} (left) shows the dimension-specific self-bias estimates. GPT, Claude and Mistral models show a consistent trend across most dimensions, exhibiting no or relatively small positive self-bias. However, we observe outliers to the overall trend of each model, such as the logical correctness dimension for GPT models and Claude-3.5-Sonnet, and the faithfulness dimension for Mistral Large, where the models show higher self-bias compared to what they do in other dimensions.

A more sharp difference is observed for the Llama models. Notably, Llama 3 8B has significantly larger negative self-bias in faithfulness. As confirmed by \Cref{fig:heatmap-scores-all-dimensions}, while human raters perceive minimal differences in faithfulness across models, the Llama models consistently assign lower scores—particularly to their own completions. This suggests these models may be excessively critical regarding their own faithfulness.

\paragraph{Analysis by task type} 
We also split the data based on the task type (open-ended QA and summarization) and analyze self-bias separately for each task category in \Cref{fig:bias-by-dimension-and-dataset} (right). Most models show higher self-bias for open-ended QA tasks than summarization, a phenomenon suggesting a relationship between task category and self-bias. However, this is not the only possibility.
The summarization group consists of older datasets (prior to 2023), than the open-ended QA. This introduces the possibility of data contamination; \llmaaj has been instructed on this data to prefer completions from humans or a teacher model, resulting on a correction of its natural tendency towards positive self-bias. Unfortunately we cannot test this hypothesis, as we need the training data and the instruction-tuning methodology for each of these models, leaving it to future work.

\subsection{Robustness Checks} 
\label{sec:robustness}
Next, we discuss the additional analyses to check whether our conclusions hold under different assumptions of the data-generating mechanism. All visualizations are deferred to \Cref{sec:appendix-additional-results}.

\paragraph{Controlling for length} It is possible that the evaluator and reference scores may differ in their preference for different completion lengths. Thus, we augment the regression in \Cref{eq:main_model} by adding a term to control for length for each judge separately. Concretely, we define a normalized length for each completion as: $\smash{\tilde{\ell}_{im}=(l_{im}-\bar{l}_i)/\sqrt{\Var(l_{i})}}$ where $l_{im}$ is the token-length of the completion using the BERT tokenizer \citep{devlin2019bert} from model $m$ on prompt $i$, while $\bar{l}_i$ denotes denote the average length across all model completions for that prompt. Following \citet{dubois2024length}, we then take the hyperbolic tangent of $\smash{\tilde{\ell}_{im}}$ to bound the values. Normalizing length using global means and variances yields comparable results. In both cases, we observe that \llmaaj are positively correlated with length, both overall and separately for each dimension. This means that longer lengths are associated with higher ratings, which agrees with findings from previous studies \citep{saito2023verbosity}. However, once reference scores are accounted for in the regression, this association effectively disappears. 
Consistently, including the length-control term in the regression does not meaningfully affect our estimates of self- and family-bias, which remain virtually unchanged (see \Cref{fig:bias-length-control}).

\paragraph{Varying the model specification} We change the model specification in two ways. First, given that evaluation dimensions differ in the granularity of their rating scales, we replace the linear regression model from \Cref{eq:main_model} with different ordered logistic regressions and fit them separately for each dimension. 
We find that self-bias remains positive for the GPT models and for Claude 3.5-Sonnet, while the Llama models still exhibit the negative self-bias for faithfulness (\Cref{fig:bias-by-dimension-with-logits}). 
Second, we relax the assumption of linear dependence between evaluator and reference ratings by fitting generalized additive models (GAMs) that model this relationship using cubic splines \citep{hastie2017generalized}. Our main conclusions regarding self- and family-bias remain qualitatively unchanged under this alternative model specification and thus we omit the results. 

\paragraph{\llmaaj ratings as reference scores} As discussed in \Cref{sec:methods}, human judgments may not perfectly represent the  quality intended to be measured by each evaluation dimension. 
A possibility is that LLM judgments may be more accurate than humans. 
However, simply replacing human scores with LLM judge scores without adjusting the regression would be problematic, as it would introduce circularity. 
To address this issue, we proceed as follows. For each model family, we remove judgments and completions generated by models belonging to that family from the data. Then, for each remaining completion, we compute the average rating assigned by judges within the excluded family and use this average as an alternative reference score. 
Under this alternative reference scoring scheme, our estimates of self- and family-bias remain qualitatively similar to those obtained using human scores, as seen in \Cref{fig:bias-by-switching-reference}: GPT models as well as Sonnet 3.5 still exhibit strong self-bias regardless of which scores are used as reference, Llama 3 8B shows negative self-bias, while the magnitude of the self-bias for the others is small. Family bias remains substantial for GPT and Claude models. 

\paragraph{Removing the weakest models} We do another sanity check by removing the weakest models (based on the scores we have seen) from the data, namely Mistral 7B and Llama 3 8B. 
Since they obtain lower scores than other models, we need to ensure that our results are robust to their removal and thus remove their completions from the data and rerun the analysis. The estimated self- and family-biases are shown in \Cref{fig:bias-wo-small-capacity}. We observe that the magnitude of the self-bias slightly decreases for all models; this is potentially explained by the fact that, with the weakest models removed, the overall range in completion quality narrows, leaving less scope for substantial differences in how judges score their own outputs relative to others. However, GPT-4o and Claude-3.5-Sonnet's remain statistically significant. Family-bias for these families also remains positive.

\section{Conclusions \& Future Work}\label{sec:discussion}

In this work, we propose a statistical approach -- which integrates human reference scores and accounts for judge-specific effects via regression analysis—to quantify self-bias and family-bias in \llmaaj. By explicitly modeling the alignment between LLM scores and an independent annotator, our approach isolates the systematic favoritism where models rate their own outputs, as well as outputs from other models within the same family, more highly than warranted by true performance differences. Our analysis shows that models like GPT-4o and Claude 3.5 Sonnet have significant self-bias in some evaluation dimensions and datasets but not in others. The extent of self-bias varies depending on the evaluation scenario. Additionally, we observed family-bias, indicating systematic favoritism among models with similar architectures, training methods, or styles.

Our findings highlight the importance of explicitly measuring and reporting self-bias in \llmaaj. If reference scores from an independent judge are available, practitioners can obtain unbiased judge scores by subtracting statistically estimated self- and family-bias from the \llmaaj ratings. This procedure yields debiased evaluation scores and can be applied directly at deployment time, ensuring consistent evaluation for new model outputs with similar characteristics. Additionally, our framework provides practical guidance for estimating unbiased reference scores on a dataset when only a limited number of human annotations—but many \llmaaj ratings—are available, through stratified prediction-powered inference \citep{fogliato2024framework, fisch2024stratified}. Specifically, we have shown that it is crucial to stratify completions according to both the evaluation dimension (e.g., correctness vs. conciseness) and whether the evaluated model belongs to the judge's family (e.g., GPT evaluating GPT outputs), as this stratification substantially reduces the variance of bias estimates. In scenarios where human annotations are entirely unavailable, assembling a diverse panel of \llmaaj judges \citep{verga2024replacing,badshah2024reference,liprd} drawn from multiple model families (such as GPT, Claude, and Llama) further minimizes systematic biases, ensuring fair, consistent, and robust evaluations across different models and evaluation tasks over time.

Our proposed approach can be applied to measure other types of biases, as long as there is a distinct control-group of completions where the bias does not apply, and a benchmark reference score mechanism, that we want to imitate, such as human annotations. Some interesting directions for future work include applying our approach to different types of bias and studying the cause and extend of negative self-bias using a white-box \llmaaj, where we know the training data and process.

\section*{Ethics statement}

\paragraph{Limitations} Our analysis assumes human ratings as unbiased reference scores, yet human annotators may introduce subjective variability. This variability can inflate self- and family-bias estimates if actual performance differences among models are not fully captured. Although robustness checks confirm that our conclusions remain stable when replacing human scores with \llmaaj, future work would benefit from developing a more objective ground-truth measure of quality and conducting a deeper analysis of subjective versus objective evaluation dimensions. 
Note that our regression model (\Cref{eq:main_model}) implicitly also assumes that systematic biases -- apart from self-bias -- are shared by human and judge ratings, such as preferences for completion length or style. 
If this assumption is violated (e.g., the judge strongly prefers longer completions while humans do not), we might incorrectly estimate this as self-bias. Specifically, if the evaluator’s own completions are shorter on average, we could mistakenly conclude that no self-bias exists, even if the evaluator is inflating ratings of its own outputs.
Additionally, converting Likert scales into numerical scores introduces another potential limitation, as differences in granularity and interpretation across evaluation dimensions could affect comparability. Our evaluation also covers a limited set of dimensions and a fixed dataset, which might not represent broader aspects of LLM behavior across other tasks or domains. Finally, while robustness checks address some confounders, other factors such as output style or prompt difficulty may still influence bias measurements.

\paragraph{Impact} We analyze self-bias in LLM-based evaluators using anonymized, publicly available data and transparent statistical methods. We recognize that both human annotations and training data can contain inherent biases, which may influence our findings. By quantifying self- and family-bias, our work aims to inform and mitigate potential unfairness in automated evaluations. We caution that deploying LLMs as evaluators without addressing these biases could perpetuate systemic issues. Our study is presented with full disclosure of limitations, and we encourage ongoing scrutiny and improvement in ethical AI practices.

\section*{Acknowledgments}

The authors were members of AWS Bedrock at the time of writing this paper and multiple resources and people facilitated this research. We used the prompt templates co-developed with other members of the AWS Bedrock Evaluation Science team during the development of our LLM-as-a-judge product. We thank Ling Liu and Yang Li for their contribution to the prompt engineering effort. We thank the Bedrock Data team for developing and testing the human evaluation guidelines through multiple iterations of human data collections. We also thank our internal data annotation team (BDB team) led by Madhu Koneru for their high quality data annotations used in this paper. Finally, we thank Swabha Swayamdipta and Artidoro Pagnoni for their feedback to improve the manuscript.

\bibliographystyle{plainnat}
\bibliography{custom}

\appendix

\clearpage
\appendix

\section{Additional Details on the Methods}\label{sec:appendix-methods}
We provide some details on the estimation of the regression coefficients for statistics-savvy readers. Let
$\hat\gamma$ be the OLS estimate of $\smash{\gamma = (\gamma_1, \dots, \gamma_J)}$.
Under standard assumptions \citep{wooldridge2010econometric}, as
$N\rightarrow\infty$ we have
$\smash{\widehat{\Var}(\hat\gamma_j)^{-1/2} \bigl(\hat\gamma_j-\gamma_j\bigr) \stackrel{d}{\to} N(0,1)}$,
where $\widehat{\Var}(\hat\gamma_j)^{1/2}$ is a consistent estimator of the
White standard error of $\hat\gamma_e$ \citep{kuchibhotla2018model, fogliato2021maars}.  
Assessing the presence of self-bias for judge $j$ boils down to testing the following (simple) null
hypothesis against its alternative:

$H_0: \gamma_j = 0$ vs. $H_1: \gamma_j \neq 0$. In other words, 
we assess whether the coefficient $\gamma_j$ is equal to $0$. 
The process is analogous for the cofficient corresponding to family-bias. 
A two-sided Wald test of level $\alpha$ will reject $H_0$ if
    $|\widehat{\Var}(\hat\gamma_j)^{-1/2} \hat\gamma_j| > z_{1-\alpha/2}$
where $z_{1-\alpha/2}$ is the $\smash{1-\alpha/2}$ percentile of a standard
Normal. 
The corresponding $\smash{(1-\alpha)}$ confidence interval is
$\smash{\hat\gamma_j \pm z_{1-\alpha/2}\,\widehat{\Var}(\hat\gamma_j)^{1/2}}$.

\section{Additional Details on the Data Collection}\label{sec:appendix-human_quality}

\paragraph{Evaluation dimensions}
\Cref{tab:quality_dimensions} shows examples that should receive high and low scores for each evaluation dimension. \Cref{tab:metric_details} shows the evaluation dimensions and associated Likert scales, as shown to annotators. 

\paragraph{Attention checks}
Attention check items are prompt-completion pairs that are deliberately created to earn low ratings on particular dimensions. These included completions that were randomly paired with other prompts from the dataset, completions in which the word order within each sentence was reversed, and completions in which up to a third of characters were removed. During the annotation collection process, any annotator who repeatedly failed attention checks was removed from the task and their annotations were re-worked by other annotators. In the final dataset, average accuracy on attention checks was over 97\%.

\paragraph{Human agreement on the full dataset}
For each metric, we compute chance-corrected inter-rater reliability using Krippendorff's $\alpha$. Due to known problems with chance-corrected measures of inter-rater reliability when applied to datasets with highly skewed label distributions \citep{zhao2013assumptions}, we additionally present observed agreement. \Cref{tab:quality} shows both measurements of inter-rater reliability for the different dimensions. 

\paragraph{Human agreement on the gold dataset}
Three passes of gold annotations were obtained from a separate team of expert annotators. The gold annotations were aggregated by taking the lowest and highest values of the three, resulting in a gold range. Any annotation from our originial dataset which falls within the boundaries of the gold range was considered accurate. We additionally present correlation values between average ratings for the two annotator populations, using Spearman's $\rho$. \Cref{tab:quality} shows the accuracy and correlation values for the different dimensions.

\begin{table}[t]
\begin{tabular}{|p{1.9cm}|p{4.5cm}|p{1cm}|p{5cm}|}
\hline
\textbf{Dimension} & \textbf{Input} & \textbf{Quality} & \textbf{Output} \\
\hline
\hline
\multirow{2}{1.8cm}{\textbf{Completeness}} & 
\multirow{2}{4.5cm}{Give me a list of 10 restaurants in North Platsville with a yelp rating of 5.} & High &
There are only 7 restaurants with a yelp rating of 5 in North Platsville. Here they are: [list of 7 restaurants] \\ \cline{3-4}
& & Low &
There are only 7 restaurants with a yelp rating of 5 in North Platsville. \\
\hline
\hline
\multirow{2}{1.8cm}{\textbf{Conciseness}} & 
\multirow{2}{4.5cm}{Who was president of the US in 1954?} & High &
Dwight Eisenhower was president of the US in 1954. \\ \cline{3-4}
& & Low &
Adlai Stevenson lost the presidential race in 1954 \\
\hline
\hline
\multirow{2}{1.8cm}{\textbf{Logical Robustness}} & 
\multirow{2}{4.5cm}{Please give me a brief, formal answer. Is a kilo of feathers heavier than a pound of steel?} & High &
One pound is equal to about 0.45 kilograms. Therefore, one kilogram of feathers is heavier than one pound of steel. \\ \cline{3-4}
& & Low &
One pound equals 0.45 kilograms... [contradictory explanation about weights and units] \\
\hline
\hline
\multirow{2}{1.8cm}{\textbf{Logical Correctness}} & 
\multirow{2}{4.5cm}{[...] The Norman economy struggled... What was one of the Norman's major exports?} & High &
The Normans were known for exporting fine pottery. \\ \cline{3-4}
& & Low &
The Normans were only known for exporting racing donkeys. \\
\hline
\hline
\multirow{2}{1.8cm}{\textbf{Helpfulness}} & 
\multirow{2}{4.5cm}{Mara has six apples... How many apples does Mara think Myra has?} & High &
Based on the information provided... [detailed explanation about apple scenario] \\ \cline{3-4}
& & Low &
Three. Two. One. \\
\hline
\hline
\multirow{2}{1.8cm}{\textbf{Faithfulness}} & 
\multirow{2}{4.5cm}{Kevin Presto is requesting... What is this email about?} & High &
The email is about a meeting on Wednesday, January 2 at 9:30 am \\ \cline{3-4}
& & Low &
The email is about a meeting on Wednesday, January 3 at 9:30 am \\
\hline

\end{tabular}
\caption{Quality dimensions with high and low rating examples.}
\label{tab:quality_dimensions}
\end{table}

\begin{table}[t]
\centering
\begin{minipage}{\textwidth}
\centering
\adjustbox{max width=\textwidth, max totalheight=0.97\textheight}{
\begin{tabular}{m{0.15\linewidth} m{0.25\linewidth} m{0.6\linewidth}}
\toprule
\textbf{Evaluation Dimension} & \textbf{Question Shown to Annotators} & \textbf{Rating Options} \\
\midrule
\multirow{5}{=}{Completeness} & \multirow{5}{=}{Does the Output contain the necessary amount of information and detail for answering the Input?} & Not at all: none of the necessary information and detail is present. \\
\cmidrule(l){3-3}
 &  & Not generally: less than half of the necessary information and detail is present. \\
\cmidrule(l){3-3}
 &  & Neutral/mixed: about half of the necessary information and detail is present, or it's unclear what the right amount of information is. \\
\cmidrule(l){3-3}
 &  & Generally yes: most of the necessary information and detail is present. \\
\cmidrule(l){3-3}
 &  & Yes: all necessary information and detail is present. \\
\midrule
\multirow{5}{=}{Conciseness} & \multirow{5}{=}{How focused is the Output on the Input?} & Not at all: no part of the output is focused on the input. \\
\cmidrule(l){3-3}
 &  & Slightly: an overwhelming amount of the output is irrelevant or the relevant information is not a direct answer. \\
\cmidrule(l){3-3}
 &  & Somewhat: roughly half of the output is relevant to the input. \\
\cmidrule(l){3-3}
 &  & Mostly: an overwhelming amount of the output is relevant to the input. \\
\cmidrule(l){3-3}
 &  & Completely: every piece of the output is relevant to the input. \\
\midrule
\multirow{5}{=}{Logical robustness} & \multirow{5}{=}{Do the arguments presented in the Output follow logically from one another?} & Not at all: the Output contains too many errors of reasoning to be usable. \\
\cmidrule(l){3-3}
 &  & Not generally: the output contains a few instances of coherent reasoning, but errors reduce the quality of the Output. \\
\cmidrule(l){3-3}
 &  & Neutral/mixed: I can't tell if the reasoning is correct -- different users may disagree. \\
\cmidrule(l){3-3}
 &  & Generally yes: the Output contains small issues with reasoning but the main point is supported. \\
\cmidrule(l){3-3}
 &  & Yes, completely: there are no issues with logical robustness at all. \\
\midrule
\multirow{5}{=}{Logical correctness} & \multirow{5}{=}{Is the Output a correct and accurate response to the Input?} & The response is clearly incorrect. \\
\cmidrule(l){3-3}
 &  & The response partially correct. \\
\cmidrule(l){3-3}
 &  & The response is completely correct. \\
\cmidrule(l){3-3}
 &  & NA: not enough information to determine Correctness. \\
\cmidrule(l){3-3}
 &  & NA: the Input does not expect a definitively correct answer. \\
\midrule
\multirow{7}{=}{Helpfulness} & \multirow{7}{=}{How helpful would most users find this Output?} & Not helpful at all. \\
\cmidrule(l){3-3}
 &  & Very unhelpful. \\
\cmidrule(l){3-3}
 &  & Somewhat unhelpful. \\
\cmidrule(l){3-3}
 &  & Neutral/Mixed. \\
\cmidrule(l){3-3}
 &  & Somewhat helpful. \\
\cmidrule(l){3-3}
 &  & Very helpful. \\
\cmidrule(l){3-3}
 &  & Above and beyond. \\
\midrule
\multirow{6}{=}{Faithfulness} & \multirow{6}{=}{How much of the information in the Output is contained in the Input or Retrieved Passages (or can be easily inferred from these sources via common sense knowledge)?} & Not at all: none of the information in the output is contained in the input or retrieved passages. \\
\cmidrule(l){3-3}
 &  & Not generally: some of the information in the output is contained in the input or retrieved passages. \\
\cmidrule(l){3-3}
 &  & Neutral/mixed: approximately half of the information in the output is contained in the input or retrieved passages. \\
\cmidrule(l){3-3}
 &  & Generally yes: most of the information in the output is contained in the input or retrieved passages. \\
\cmidrule(l){3-3}
 &  & Yes: all of the information in the output is contained in the input or retrieved passages. \\
\cmidrule(l){3-3}
 &  & NA: the request does not expect the model to stay faithful to a specific piece of text in the context. \\
\bottomrule
\end{tabular}
}
\end{minipage}
\caption{Scoring rubric shown to annotators for evaluation dimensions.}
\label{tab:metric_details}
\end{table}

\begin{table}[H]
    \centering
    \begin{tabular}{m{0.2\linewidth} m{0.15\linewidth} m{0.15\linewidth} m{0.15\linewidth} m{0.18\linewidth}}
        \toprule
        \multirow{2}{=}{\textbf{Evaluation dimension}} & \multicolumn{2}{c}{\textbf{Gold subset}} & \multicolumn{2}{c}{\textbf{Full dataset}} \\
        \cmidrule(lr){2-3} \cmidrule(lr){4-5}
         & \textbf{Accuracy} & \textbf{Correlation} & \textbf{Agreement} & \textbf{Kripendorff's $\alpha$}\\
        \midrule
        Completeness        & 0.87 & 0.47 & 0.67 & 0.47 \\
        Conciseness         & 0.95 & 0.12 & 0.88 & 0.15 \\
        Logical Robustness  & 0.95 & 0.23 & 0.87 & 0.14 \\
        Logical Correctness & 0.92 & 0.37 & 0.90 & 0.31 \\
        Helpfulness         & 0.84 & 0.56 & 0.68 & 0.47 \\
        Faithfulness          & 0.92 & 0.21 & 0.83 & 0.15 \\
        \midrule
        Average             & 0.91 & 0.33 & 0.81 & 0.28 \\
        \bottomrule
    \end{tabular}
    \caption{Quality assessment of human annotations per dimension.}\label{tab:quality}
\end{table}

\clearpage
\paragraph{Completions and \llmaaj judgments} All model completions and \llmaaj judgments were obtained on November 2024, by calling the corresponding APIs. All models were prompted in an identical way (no prompt engineering). The prompt templates for \llmaaj for each dimension are provided in \Cref{sec:prompt_templates}.

\begin{figure}
    \centering
    \includegraphics[width=0.9\linewidth]{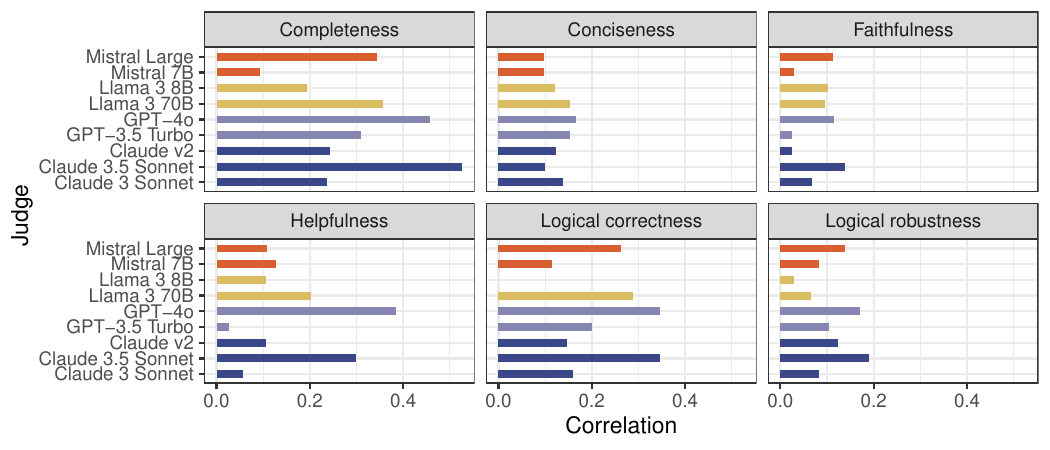}
    \caption{Tie-corrected Spearman rank correlation between \llmaaj and humans.}
    \label{fig:llmaaj_human_spearman}
\end{figure}

\paragraph{\llmaaj correlation with humans} \Cref{fig:llmaaj_human_spearman} shows the tie-corrected Spearman correlation of each \llmaaj with humans. We observe higher correlations in dimensions with high human inter-annotator agreement. This phenomenon is partially due to the highly imbalanced classes observed in dimensions such as conciseness and logical robustness. 

\clearpage 
\section{Additional Results}
\label{sec:appendix-additional-results}

\subsection{Additional Visualizations}

Here we present additional visualizations that complement the results in the main body of the paper. 

\begin{figure}[ht]
    \centering
    \includegraphics[width=0.8\textwidth]{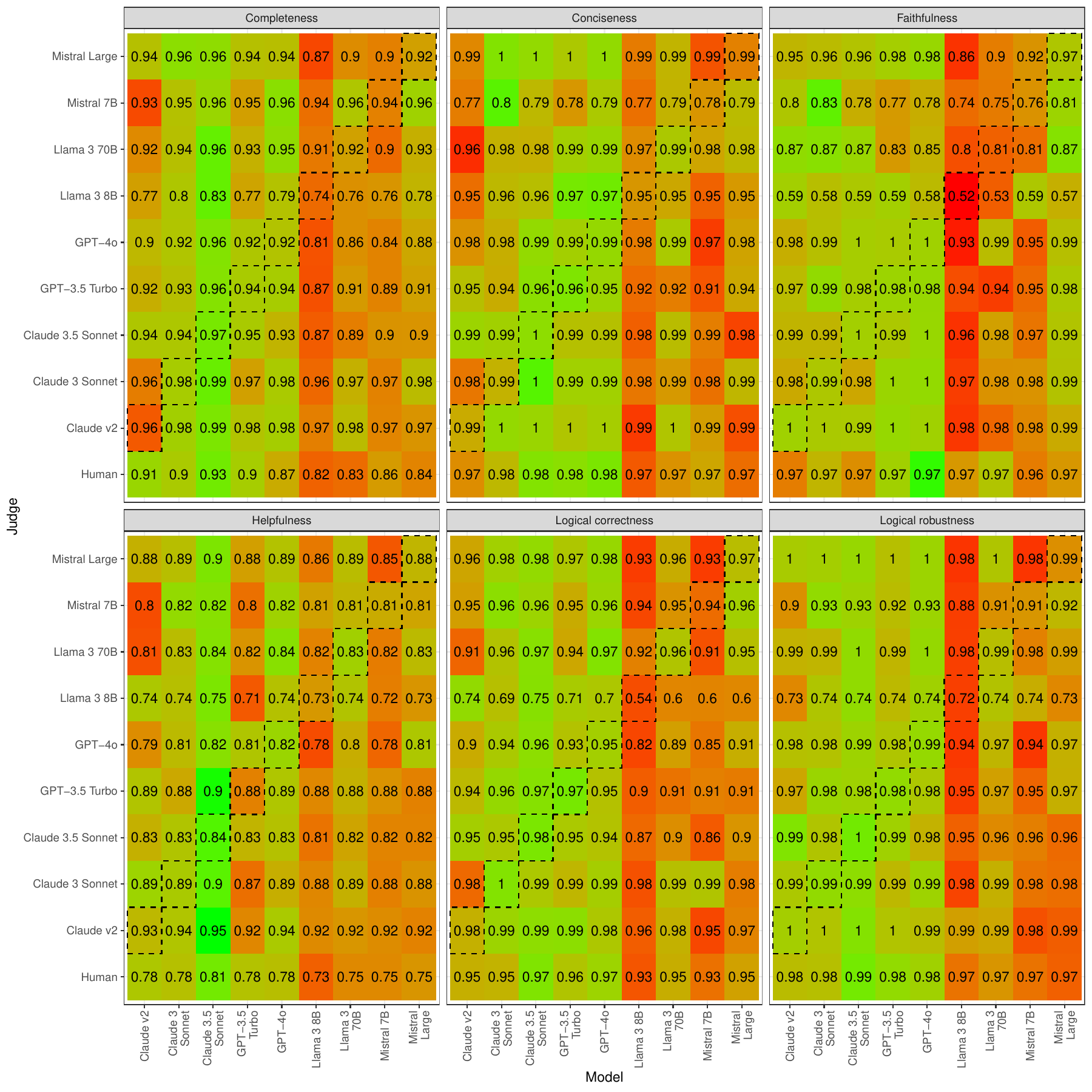}
    \caption{Heatmap of average ratings of model completions by dimension.}
    \label{fig:heatmap-scores-all-dimensions}
\end{figure}

\begin{figure}[ht]
    \centering
    \includegraphics[width=\textwidth]{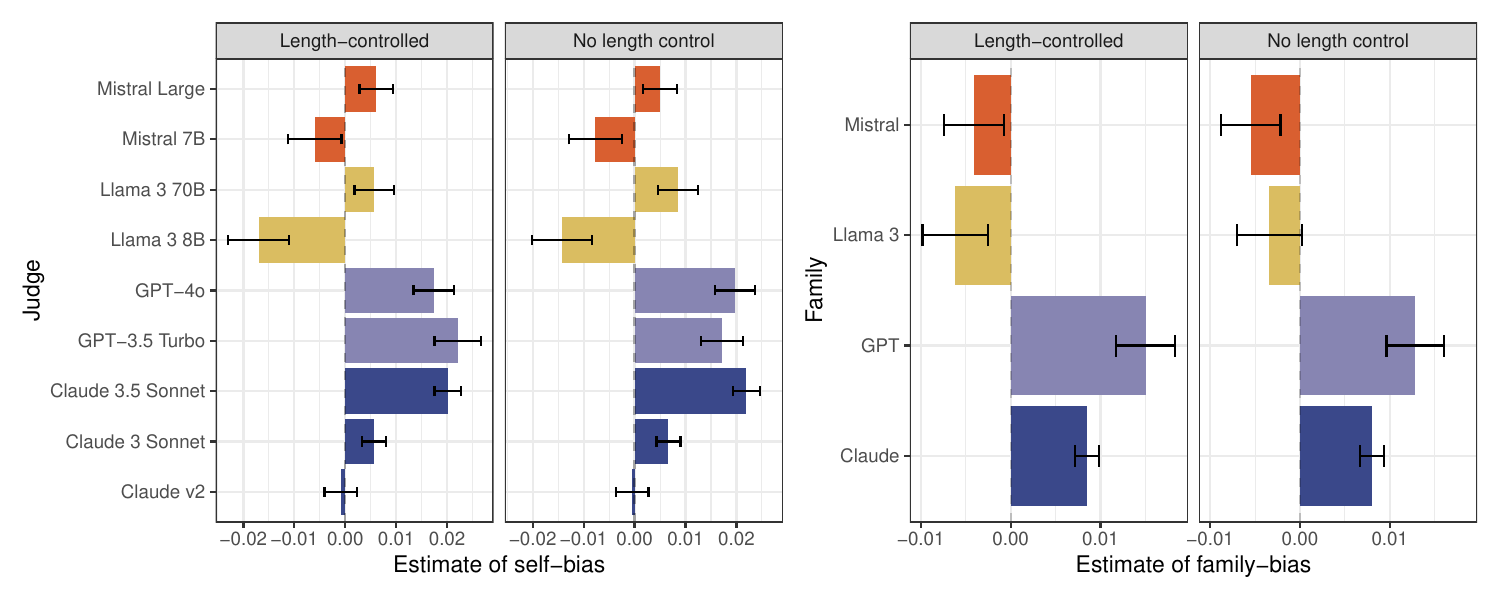}
    \caption{Robustness check: Estimates of self-bias (left) and family-bias (right) with and without length control. Results without length control correspond to \cref{fig:self-and-family-bias}.}
    \label{fig:bias-length-control}
\end{figure}

\begin{figure}[ht]
    \centering
    \includegraphics[width=0.6\textwidth]{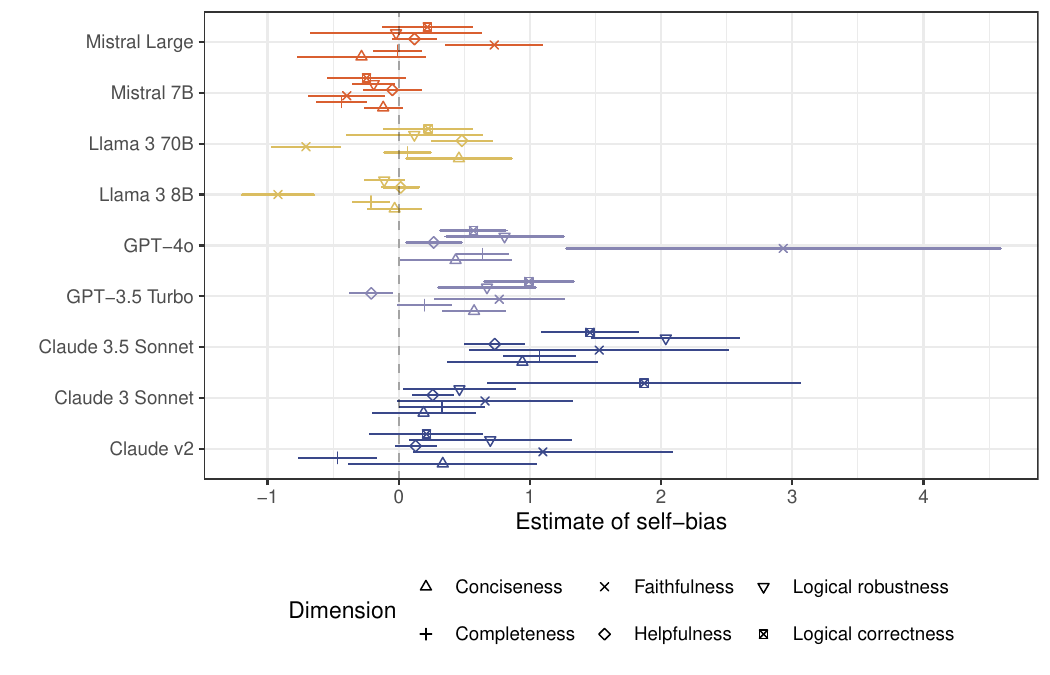}
    \caption{Robustness check: Estimates of self-bias (left) and family-bias (right) for each dimension, obtained using a logit link in \cref{eq:main_model}. }
    \label{fig:bias-by-dimension-with-logits}
\end{figure}

\begin{figure}[t]
    \centering
    \includegraphics[width=\textwidth]{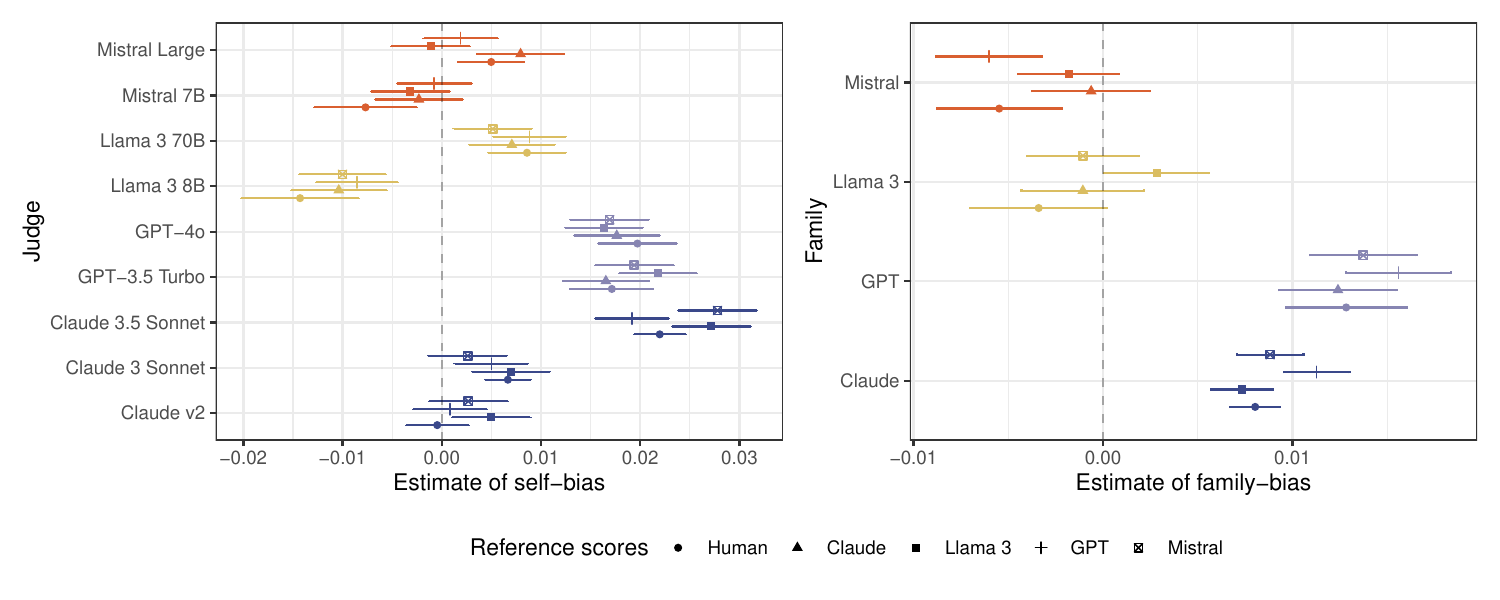}
    \caption{Robustness check: Estimates of self-bias (left) and family-bias (right) obtained using different reference scores.}
    \label{fig:bias-by-switching-reference}
\end{figure}

\begin{figure}[t]
    \centering
    \includegraphics[width=\textwidth]{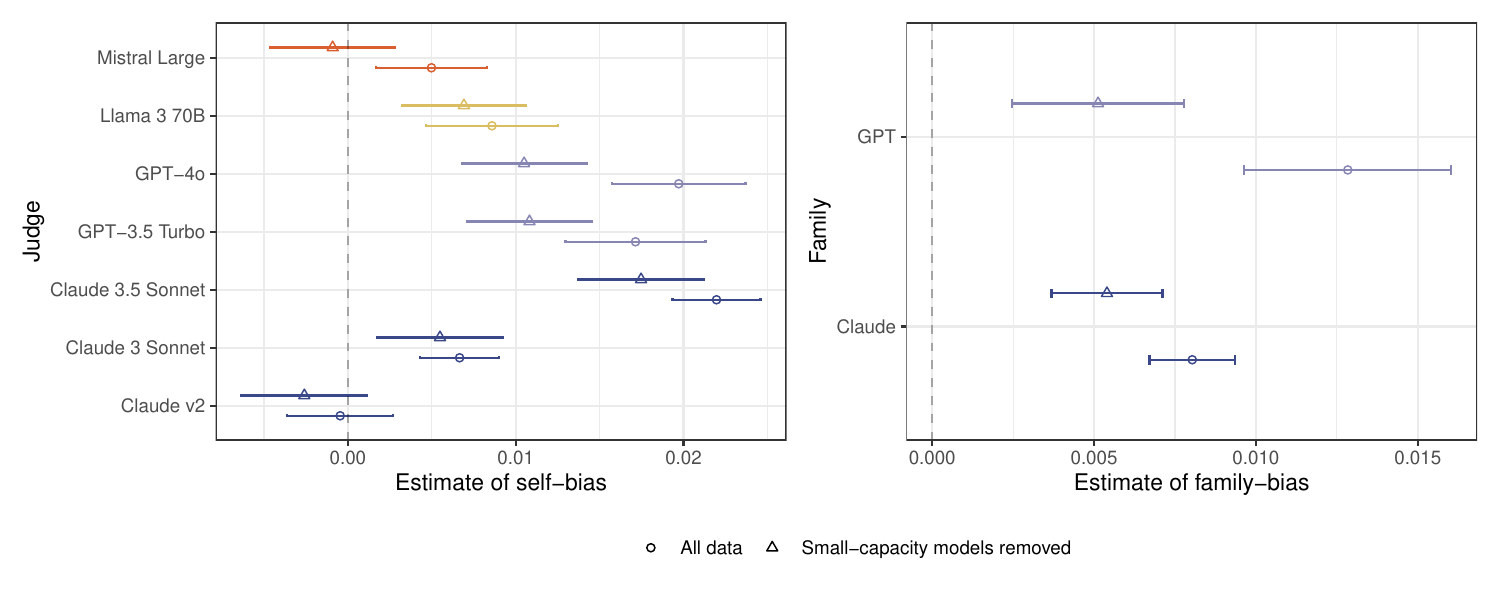}
    \caption{Robustness checks: Estimates of self-bias (left) and family-bias (right) obtained with and without small capacity models (Claude v2, Llama 3 8B, and Mistral 7B) in the data.}
    \label{fig:bias-wo-small-capacity}
\end{figure}

\clearpage
\subsection{Additional Results on HELM Instruct}\label{sec:helm-instruct-analysis}

We additionally conduct an analysis using data from HELM Instruct \citep{helm-instruct}, a dataset that in part we also use in our main study. The dataset consists of open-ended prompts drawn from diverse instruction-following scenarios, including dialogues, question answering, and general-purpose tasks. Each model response is evaluated by crowdworkers along five criteria—Helpfulness, Understandability, Completeness, Conciseness, and Harmlessness—on a 1-to-5 scale. 

The dataset contains model completions from four instruction-following LLMs: GPT-4 (0314), GPT-3.5 Turbo (0613), Anthropic Claude v1.3, and Cohere-Command-Xlarge-Beta. Each model was evaluated using judgments from both human annotators, collected via Amazon Mechanical Turk, and two LLM-based evaluators: GPT-4 (0314) and Claude v1.3. We use the MTurk human ratings as reference scores throughout our analysis, treating them as independent judgments against which model evaluation behavior—including self-bias—can be compared.

\begin{figure}[t]
    \centering
    \includegraphics[width=\textwidth]{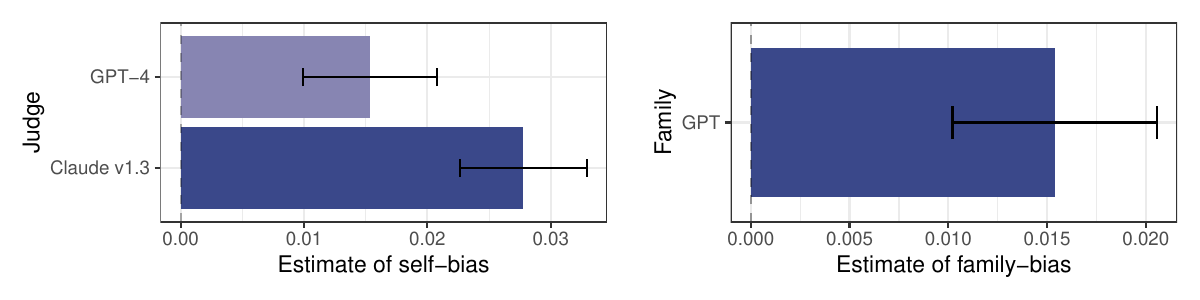}
    \caption{HELM-Instruct: Estimates of self-bias (left) and family-bias (right) for HELM-Instruct data and judges.}
    \label{fig:bias-helm}
\end{figure}

\Cref{fig:bias-helm} shows estimated self- and family-bias coefficients for Claude v1.3 and GPT-4, as evaluators on the HELM Instruct dataset. We observe a positive self-bias for both models, with GPT-4 showing a slightly larger magnitude. This indicates that both models tend to assign higher scores to their own completions, even after accounting for completion quality via human reference scores. Additionally, GPT-4 shows family bias.

\clearpage 
\section{Prompt Templates}
\label{sec:prompt_templates}

Below we present the prompt templates used for the \llmaaj. The same prompts were used across all models. 


\begin{tcolorbox}[colback=gray!5!white,colframe=black!75!white,title=Faithfulness Prompt]
You are given a task in some context (\textbf{Input}), and a candidate answer. Is the candidate answer faithful to the task description and context?

A response is considered \textit{unfaithful} only when (1) it clearly contradicts the context, or (2) the task implies that the response must be based on the context (e.g., a summarization task). If the task does not require grounding in the context, the model may use its own knowledge, even if unverifiable.

\vspace{1mm}
\noindent\textbf{Task:} \{prompt\}\\
\textbf{Candidate Response:} \{prediction\}

\vspace{1mm}
\noindent\textbf{Instruction:} Evaluate how much of the information in the answer is faithful to the available context.

First explain your reasoning, then provide your final answer. Use the following format:

\begin{quote}
Explanation: [Explanation], Answer: [Answer]
\end{quote}

\noindent where \texttt{[Answer]} is one of:
\begin{quote}
\texttt{none is faithful}\\
\texttt{some is faithful}\\
\texttt{approximately half is faithful}\\
\texttt{most is faithful}\\
\texttt{all is faithful}
\end{quote}
\end{tcolorbox}

\begin{tcolorbox}[colback=gray!5!white,colframe=black!75!white,title=Logical Robustness Prompt]
You are given a task in some context (\textbf{Input}), and a candidate answer. Evaluate whether the arguments in the response follow logically from one another.

Consider the following aspects:
\begin{itemize}
\item Self-contradictions within the response
\item Logic gaps or errors in reasoning
\item Soundness of reasoning (given the premises)
\item Proper argumentation where required
\end{itemize}

Note that factual correctness is separate from logical cohesion - evaluate the reasoning process, not the accuracy of claims.

\vspace{1mm}
\noindent\textbf{Task:} \{prompt\}\\
\textbf{Candidate Response:} \{prediction\}

\vspace{1mm}
\noindent\textbf{Instruction:} Evaluate the logical cohesion of the response.

First explain your reasoning, then provide your final answer. Use the following format:

\begin{quote}
Explanation: [Explanation], Answer: [Answer]
\end{quote}

\noindent where \texttt{[Answer]} is one of:
\begin{quote}
\texttt{Not at all}\\
\texttt{Not generally}\\
\texttt{Neutral/Mixed}\\
\texttt{Generally yes}\\
\texttt{Yes}
\end{quote}
\end{tcolorbox}

\begin{tcolorbox}[colback=gray!5!white,colframe=black!75!white,title=Logical Correctness Prompt]
You are given a task in some context (\textbf{Input}), and a candidate answer. Evaluate whether the response is correct and accurate, focusing only on content and solution validity.

Note that style, presentation, format, or language issues should not affect the evaluation of correctness.

\vspace{1mm}
\noindent\textbf{Task:} \{prompt\}\\
\textbf{Candidate Response:} \{prediction\}

\vspace{1mm}
\noindent\textbf{Instruction:} Evaluate whether the response is correct and accurate for the given task.

First explain your reasoning, then provide your final answer. Use the following format:

\begin{quote}
Explanation: [Explanation], Answer: [Answer]
\end{quote}

\noindent where \texttt{[Answer]} is one of:
\begin{quote}
\texttt{correct}\\
\texttt{partially correct}\\
\texttt{incorrect}
\end{quote}
\end{tcolorbox}

\begin{tcolorbox}[colback=gray!5!white,colframe=black!75!white,title=Helpfulness Prompt]
You are given a task in some context (\textbf{Input}), and a candidate answer. Evaluate how helpful the completion is for the user's request.

A response is considered \textit{helpful} when it satisfies both explicit and implicit expectations in the user's request. Consider factors such as:
\begin{itemize}
\item Coherence and clarity given the context
\item Task completion (if applicable)
\item Following provided instructions
\item Appropriate style and format
\item Audience appropriateness
\item Specificity level
\item Conciseness vs. elaboration as needed
\item Avoiding unnecessary content
\item Anticipating user needs
\item Interest level (when appropriate)
\item Solution elegance (for technical problems)
\item Appropriate chat formatting (for conversations)
\end{itemize}

\vspace{1mm}
\noindent\textbf{Task:} \{prompt\}\\
\textbf{Candidate Response:} \{prediction\}

\vspace{1mm}
\noindent\textbf{Instruction:} Evaluate how helpful the response is for the given task.

First explain your reasoning, then provide your final answer. Use the following format:

\begin{quote}
Explanation: [Explanation], Answer: [Answer]
\end{quote}

\noindent where \texttt{[Answer]} is one of:
\begin{quote}
\texttt{above and beyond}\\
\texttt{very helpful}\\
\texttt{somewhat helpful}\\
\texttt{neither helpful nor unhelpful}\\
\texttt{somewhat unhelpful}\\
\texttt{very unhelpful}\\
\texttt{not helpful at all}
\end{quote}
\end{tcolorbox}

\begin{tcolorbox}[colback=gray!5!white,colframe=black!75!white,title=Completeness Prompt]
You are given a task in some context (\textbf{Input}), and a candidate answer. Determine whether the response contains all necessary information and detail to properly answer the input.

Focus only on information completeness, not on accuracy, style, or coherence. A response is considered \textit{incomplete} when it:
\begin{itemize}
\item Misses explicitly requested items
\item Fails to address all parts of multi-part requests
\item Provides insufficient detail
\item Misunderstands or ignores the input
\end{itemize}

For evasive responses ("I can't answer that"), rate as complete if appropriate, or evaluate the provided portion if partially evasive.

\vspace{1mm}
\noindent\textbf{Task:} \{prompt\}\\
\textbf{Candidate Response:} \{prediction\}

\vspace{1mm}
\noindent\textbf{Instruction:} Evaluate how complete the response is relative to the task requirements.

First explain your reasoning, then provide your final answer. Use the following format:

\begin{quote}
Explanation: [Explanation], Answer: [Answer]
\end{quote}

\noindent where \texttt{[Answer]} is one of:
\begin{quote}
\texttt{Not at all}\\
\texttt{Not generally}\\
\texttt{Neutral/Mixed}\\
\texttt{Generally yes}\\
\texttt{Yes}
\end{quote}
\end{tcolorbox}

\begin{tcolorbox}[colback=gray!5!white,colframe=black!75!white,title=Conciseness Prompt]
You are given a task in some context (\textbf{Input}), and a candidate answer. Assess how focused and relevant the response is to the given question.

Note that responses indicating inability to answer (e.g., "I don't know") are considered relevant if appropriate. However, irrelevant additional content should be penalized even if preceded by such statements.

\vspace{1mm}
\noindent\textbf{Task:} \{prompt\}\\
\textbf{Candidate Response:} \{prediction\}

\vspace{1mm}
\noindent\textbf{Instruction:} Evaluate how relevant and focused the response is to the task.

First explain your reasoning, then provide your final answer. Use the following format:

\begin{quote}
Explanation: [Explanation], Answer: [Answer]
\end{quote}

\noindent where \texttt{[Answer]} is one of:
\begin{quote}
\texttt{not at all}\\
\texttt{slightly}\\
\texttt{somewhat}\\
\texttt{mostly}\\
\texttt{completely}
\end{quote}
\end{tcolorbox}

\end{document}